\documentclass[10pt,twocolumn,letterpaper]{article}

\usepackage{wacv}
\usepackage{times}
\usepackage{epsfig}
\usepackage{graphicx}
\usepackage{amsmath}
\usepackage{amssymb}

\usepackage{epsfig}
\usepackage{graphicx}
\usepackage{amssymb}
\usepackage{epsfig}
\usepackage{amsmath}
\usepackage{multirow}
\usepackage{booktabs}
\usepackage[ruled]{algorithm2e}
\usepackage{subcaption}

\SetKwInOut{Parameter}{parameter}
\SetKwData{Left}{left}
\SetKwData{This}{this}
\SetKwData{Up}{up}
\SetKwFunction{Union}{Union}
\SetKwFunction{FindCompress}{FindCompress}
\SetKwInOut{Input}{input}
\SetKwInOut{Output}{output}
\SetKwInput{KwInput}{Input}                % Set the Input
\SetKwInput{KwOutput}{Output}              % set the Output
\usepackage[noend]{algpseudocode}
\makeatletter
\def\BState{\State\hskip-\ALG@thistlm}
\makeatother

\def\wacvPaperID{0190} % Enter the WACV Paper ID here

\wacvfinalcopy % *** Uncomment this line for the final submission

\ifwacvfinal
\def\assignedStartPage{9876} % *** Enter the assigned starting page number (instead of 9876)
\fi

\ifwacvfinal
\usepackage[breaklinks=true,bookmarks=false]{hyperref}
\else
\usepackage[pagebackref=true,breaklinks=true,colorlinks,bookmarks=false]{hyperref}
\fi

\ifwacvfinal
\else
\fi

\begin{document}

%%%%%%%%% TITLE
\title{Domain Impression: A Source Data Free Domain Adaptation Method}

\author{Vinod K Kurmi\\
IIT Kanpur\\
{\tt\small vinodkk@iitk.ac.in}
\and
Venkatesh K Subramanian\\
IIT Kanpur\\
{\tt\small venkats@iitk.ac.in}
\and
Vinay P Namboodiri\\
University of Bath\\
{\tt\small vpn22@bath.ac.uk}
}

\maketitle
%\thispagestyle{empty}

%%%%%%%%% ABSTRACT
\begin{abstract}
   Unsupervised Domain adaptation methods solve the adaptation problem for an unlabeled target set, assuming that the source dataset is available with all labels. However, the availability of actual source samples is not always possible in practical cases. It could be due to memory constraints, privacy concerns, and challenges in sharing data. This practical scenario creates a bottleneck in the domain adaptation problem. This paper addresses this challenging scenario by proposing a domain adaptation technique that does not need any source data. Instead of the source data, we are only provided with a classifier that is trained on the source data. Our proposed approach is based on a generative framework, where the trained classifier is used for generating samples from the source classes. We learn the joint distribution of data by using the energy-based modeling of the trained classifier. At the same time, a new classifier is also adapted for the target domain. We perform various ablation analysis under different experimental setups and demonstrate that the proposed approach achieves better results than the baseline models in this extremely novel scenario.
\end{abstract}

%%%%%%%%% BODY TEXT
\section{Introduction}
Deep learning models have been widely accepted in most of the computer vision tasks. These models, however, suffer from the problem of generalization due to dataset biases. As a result, a model trained on one dataset often performs poorly on other datasets~\cite{torralba_CVPR2011}. Domain adaptation methods try to resolve these issues by minimizing the discrepancy between the two domains. One possible way to minimize the discrepancy is by obtaining domain invariant features. These features are such that the classifier trained on one domain performs equally well on the other domains.  Domain invariant features are obtained by introducing some auxiliary tasks to minimize the distribution discrepancy of domains. To train the auxiliary task, all existing domain adaptation approaches require access to the source datasets.  The source and target datasets should both be available during the adaptation process. Nevertheless, this is not always possible in several practical scenarios. The reasons could be memory storage requirements, challenges in sharing data, privacy concerns, and other dataset handling issues. For example, the popular dataset, like Image-Net, consists of nearly 14 million images requiring hundreds of gigabytes for storage.  Another concern is related to the privacy of the dataset. In some cases, the sensitive dataset can not be shared to adapt the model for a new dataset. These limitations of the traditional domain adaptation models create a bottleneck to use it for the practical scenarios. Thus, assuming the availability of the source dataset is a severe issue in existing domain adaptation models.

\begin{figure*}
 \centering
    \includegraphics[height=4.5cm,width=14.5cm]{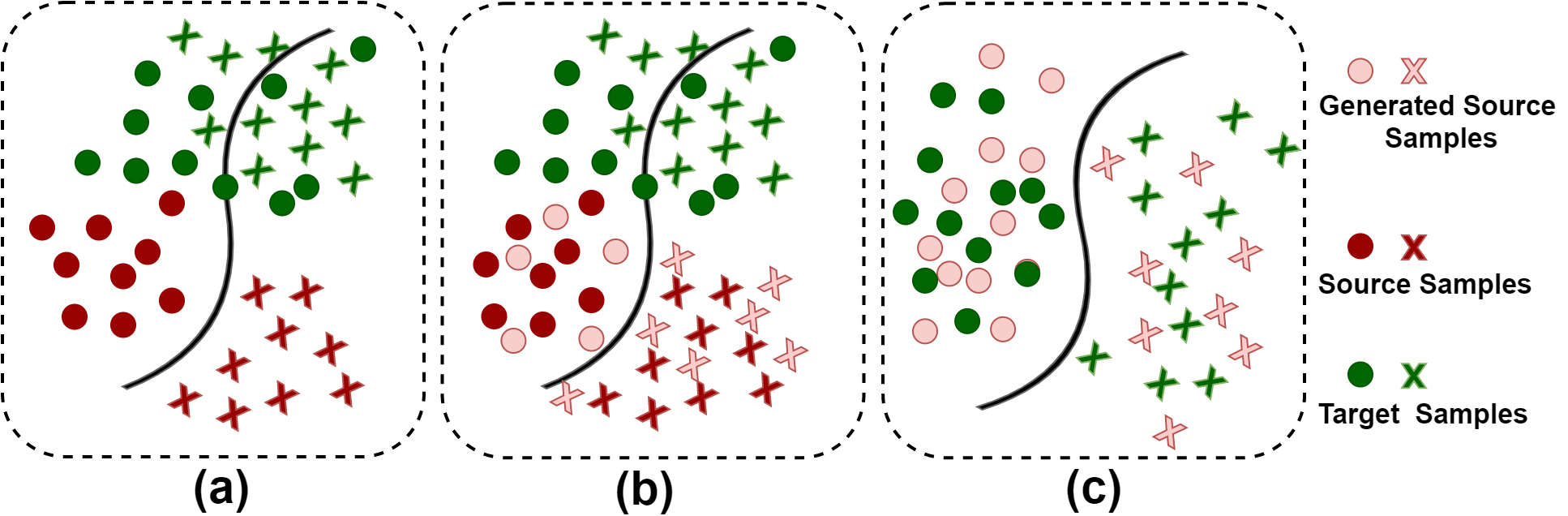}
    % \vspace{-0.5em}
       \caption{Illustration of proposed domain adaptation methods: (a) Without adaptation, the classifier trained on the source data can not correctly classify the target samples. (b) Proxy samples are generated using the trained classifier. (c) Adaptation of classifier using the proxy samples(generated). In the adaptation algorithms, only the proxy samples and target samples are used, source samples are never used in the adaptation process (best viewed in color). }
      \label{fig:intro}
    %   \vspace{-1em}
 \end{figure*}

In this paper, we propose a domain adaptation model that does not require access to source datasets at all points of time. Specifically, we assume that we have access to a classifier that is trained on the source dataset. Only the accessibility of the classifiers instead of the whole dataset makes the model utility in the practical scenarios.  We utilize the pre-trained classifier via modeling it as an energy-based function to learn the joint distribution~\cite{grathwohl2019your}. We also use a generative adversarial network (GANs) to learn the underlying data distribution of the source dataset in conjunction with this pre-trained classifier. Once the generative model is trained using the pre-trained classifier,  we proceed to generate labeled data-points that can apply in the adaptation task. We thus eliminate the need for access to the source dataset during adaptation. These generated samples can be treated as a proxy samples to train the domain adaptation model. We learn a generative function from a discriminative function by modeling it as an energy-based function. The energy of it is defined with LogSumExp() values~\cite{grathwohl2019your}. Another discriminative property of the classifier can be used with cross-entropy loss to train the generative function. Thus, the proposed method fully utilizes the information of the pre-trained classifier for the adaptation.

% The idea of learning a generative model from a discriminative model (pre-trained classifier) for the domain adaptation relies on the modeling the classifier as energy-based function and energy can be defined with LogSumExp() values~\cite{grathwohl2019your}. 
Figure~\ref{fig:intro} visualized the proposed domain adaptation framework. In Figure~\ref{fig:intro}(a) shows the distributional mismatch between source and target domain while in (b) the dummy source samples are generated using the pre-trained classifier, and the last adaptation stage is shown in (c), where the classifier is adapted for the target domain using the dummy labeled samples.

The main contributions of the proposed framework are as follows:
\begin{itemize}
    \item We provide a generative framework to tackle the source data free domain adaptation problem. 
    % \vspace{-0.5em}
    \item The trained classifier is treated as an energy-based model to learn the data distribution along with a generative adversarial network.
    % \vspace{-0.5em}
    \item We show that the generated domain impression obtained using the pre-trained classifier can be applied to other existing domain adaptation methods.
    % \vspace{-0.5em}
    \item We provide detailed ablation analysis for the proposed model to demonstrating its efficacy. We also provide comparisons with the existing baselines that use full source sample information. Our method is comparable to these baselines {\it without} using the source samples.
\end{itemize}

\section{Literature Survey}
Domain adaptation has been widely studied in the literature. All the domain adaptation frameworks try to minimize the discrepancy of source and target domain~\cite{yan_CVPR2017,saito2018maximum,ganin_ICML2015,long_arxive2017conditional}.  Reconstruction has been explored as in DRCN~\cite{ghifary_ECCV2016})and it's variants are designed to deal with two tasks, viz., classification and generation simultaneously~\cite{hoffman2018cycada,sankaranarayanan2018generate}.

\noindent\textbf{Adversarial Domain Adaptation:}  
Adversarial methods for generating images (GANs)~\cite{goodfellow_NIPS2014} were proposed a few months earlier to adversarial methods for domain adaptation using a gradient reversal layer (GRL) by Ganin and Lempitsky~\cite{ganin_ICML2015}. Adversarial domain adaptation was extended by other frameworks such as  ADDA~\cite{tzeng_CVPR2017}, TADA~\cite{wang2019transferable} and CADA~\cite{kurmi2019attending}.  These methods also suffer from the mode collapse problems. To address the mode collapse problem, multi-discriminator (MADA)~\cite{pei_arxiv2018}, CD3A~\cite{kurmi2019curriculum} and other types of discriminator based methods have been proposed~\cite{kurmi2019looking,pei_arxiv2018,wang2019transferable}. Recently there are other adversarial loss based domain adaptation methods~\cite{chen2020adversarial,tang2020discriminative,10.1145/3343031.3351070} that have been proposed to solve the domain adaptation problem more efficiently. In the drop-to-adapt method~\cite{lee2019drop} leverages adversarial dropout to learn strongly discriminative features by enforcing the cluster assumption. The augmented feature-based method~\cite{volpi_cvpr2017adversarial} proposes to minimize the discrepancy between two domains.  A conditional GAN based model has been explored in~\cite{hong2018conditional} for better semantic information. A collaborative and adversarial network (CAN)~\cite{zhang2018collaborative} has been proposed through domain-collaborative and domain adversarial training of neural networks to learn domain informative features.
% \noindent\textbf{Classifier Adaptation:}
Feature adaptation alone is not sufficient for adaptation sometimes. So classifier adaptation based methods are also introduced. Transferable adversarial training (TAT)~\cite{liu2019transferable} generates transferable examples to fill in the gap between the source and target domains and adversarially trains the deep classifiers. In~\cite{wen2019bayesian}, Bayesian uncertainty between source and target classifier is matched to adapt the classifier. 

\noindent\textbf{Privacy Concerned Domain Adaptation:}
There have been works presented to preserve the privacy of data in the learning process~\cite{abadi2016deep}. Work presented in~\cite{letien2019differentially,kieu2019domain} deals with the privacy concerns of data in domain adaptation. These models transform the data into privacy-preserving domains using some metric like optimum transport~\cite{courty2017joint}. The Federated
Transfer Learning~\cite{yang2019federated,peng2019federated} promises to combine multiple source data in the private mode. All the works so far, however,  require access to the source data for adaptation. Source data free adaptation method for off-the-self classifier~\cite{nelakurthi2018source} improves the performance of the off-the-shelf tool in the target domain by accessing some of the labeled data for the target domain. Other source data free adaptation methods~\cite{chidlovskii2016domain} are also applicable where source data is absent, but again they assume access to some of the target labels. By utilizing the classifier's information,  the model can also generate samples~\cite{grathwohl2019your}.

\noindent\textbf{Adversarial Attacks:}
The adversarial learning framework is also well explored in the adversarial attacks and perturbations~\cite{kurakin2018adversarial,moosavi2016deepfool}. These methods have been further extended for obtaining the class and data impressions~\cite{mopuri2018generalizable,nayak2019zero}. The knowledge of the classifier is also used for new unseen class samples~\cite{addepalli2019degan}. A recent work~\cite{sourcefree} suggests a domain adaptation model where source data and target data never occurred together and where class boundaries are learned in the procurement stage, while adaptation occurs in the deployment stage. However, though some works aim to reduce the need for source data, no work considers the case where source data samples are not used for training, and target labels are also not available.

\noindent\textbf{Generative Models:}
The generative approaches have successfully applied in many zero-shot recognition algorithms~\cite{xian2018feature,zhang2018visual}. In~\cite{kumar2018generalized}, authors generate novel examples from seen-unseen classes using the variational encoder-decoder. Other VAE based generative frameworks have been used in~\cite{sohn2015learning,mishra2018generative}. Similarly, in~\cite{sariyildiz2019gradient}, adversarial learning has been applied in generalized zero-shot learning. Generative adversarial network~\cite{goodfellow_NIPS2014,radford2015unsupervised} are very popular due to its capabilities of generating natural images and learn the data distribution efficiently. Conditional GAN~\cite{mirza2014conditional} also applied in many application such as cross-modal~\cite{zhang2017stackgan,xu2018attngan}, image in painting~\cite{pathak2016context,yuan2019image,yu2019free} and colorization~\cite{nazeri2018image}. Very recently, work for generative data from the trained classifier is proposed in DeepInversion~\cite{yin2020dreaming}, where the statistics of the batch normalization layer are used to obtain the training data, which could enforce the constraint on the trained classifier. Similarly, the work proposed in~\cite{santurkar2019image} generated the images from a robust classifier. The robust classifiers are trained using the robust optimization objective~\cite{madry2017towards}. Other works related to data-free distillation are resented in~\cite{chen2019data}, where a student network is trained without using the data. Similarly authors of~\cite{nayak2019zero} propose the distillation in zero shot learning framework.

Recently, there are source data free adaptation has been presented. In~\cite{li2020model}, a generative model is used to h generated target-style data using clustering-based regularization loss. SHOT~\cite{liang2020we} uses information maximization and self supervised pseudo-labeling to implicitly align representations of target and source without accesing the source data.

%  Other then domain adaptation and image generation, adversarial learning has been widely applied to understand the deep learning models. There has been some also used the adversarial learning for knowledge transfer, zero shot learning etc. Recently many methods has been proposed transfer the data knowledge from the trained classifier to another model. The class impression and data impression from the classifier has been studied by~\cite{mopuri2018generalizable,nayak2019zero} for the zero shot recognition. Other work such~\cite{grathwohl2019your} also explored the analysis of trained classifier. Adversarial attacks~\cite{kurakin2018adversarial,moosavi2016deepfool} and its variant also use the classifier information to fool the model. Recently classifier knowledge used to generate zero-shot example from the GAN~\cite{addepalli2019degan}.

\section{Background: Generative model from Discriminative model}

The objective of the discriminative model it to obtain the class conditional distribution $p(y|x)$, it focuses on the classification boundaries. Here $x$ is given input and $y$ is label. The generative models learn the joint distributions  of $p(x,y)$ from the data generation process. we rewrite the log likelihood of joint distribution distribution using the Bayes theorem as~\cite{grathwohl2019your}
\begin{equation}
\log p_\theta(x,y) = \log p_\theta(x) + \log p_\theta(y|x)
\label{eq:joint}
\end{equation}
Here $\theta$ is the parameter of the model.
The class-conditional distribution $p_\theta(y|x)$ are obtained by cross-entropy loss from the trained-classifier. The log $p_\theta(x)$ can be expressed in form of energy based models~\cite{lecun2006tutorial}. We define the log $p_\theta(x)$ as energy based functions as discussed in~\cite{grathwohl2019your}. The derivative of the log-likelihood with respect to $\theta$ can be expressed as~\cite{grathwohl2019your}
% \begin{equation}
%     p_{\theta}(\mathbf{x})=\frac{\exp \left(-E_{\theta}(\mathbf{x})\right)}{Z(\theta)}
% \end{equation}
% Energy based models say that probability density can be obtained using some of the energy function.

\begin{equation}
\frac{\partial \log p_{\theta}(\mathbf{x})}{\partial \theta}=\mathbb{E}_{p_{\theta}\left(\mathbf{x}^{\prime}\right)}\left[\frac{\partial E_{\theta}\left(\mathbf{x}^{\prime}\right)}{\partial \theta}\right]-\frac{\partial E_{\theta}(\mathbf{x})}{\partial \theta}
\label{eq:derivate}
\end{equation}
Energy functions map an input $x$ to a scalar.  We define the energy function by LogSumExp(·) of the logits of the trained classifier similar to~\cite{grathwohl2019your}

\begin{equation}
E_{\theta}(\mathbf{x})=-\log\sum_{x\in P_{\theta}(x),y}\exp \left(P_{c}(\mathbf{x})[y]\right)
\end{equation}
$P_{c}(\mathbf{x})[y]$ indicates $y^{th}$ index of output of classifier $P_{c}(\mathbf{x})$.

\section{Source Data Free Adaptation}

\begin{figure*}
 \centering
    \includegraphics[height=5.9cm,width=14.5cm]{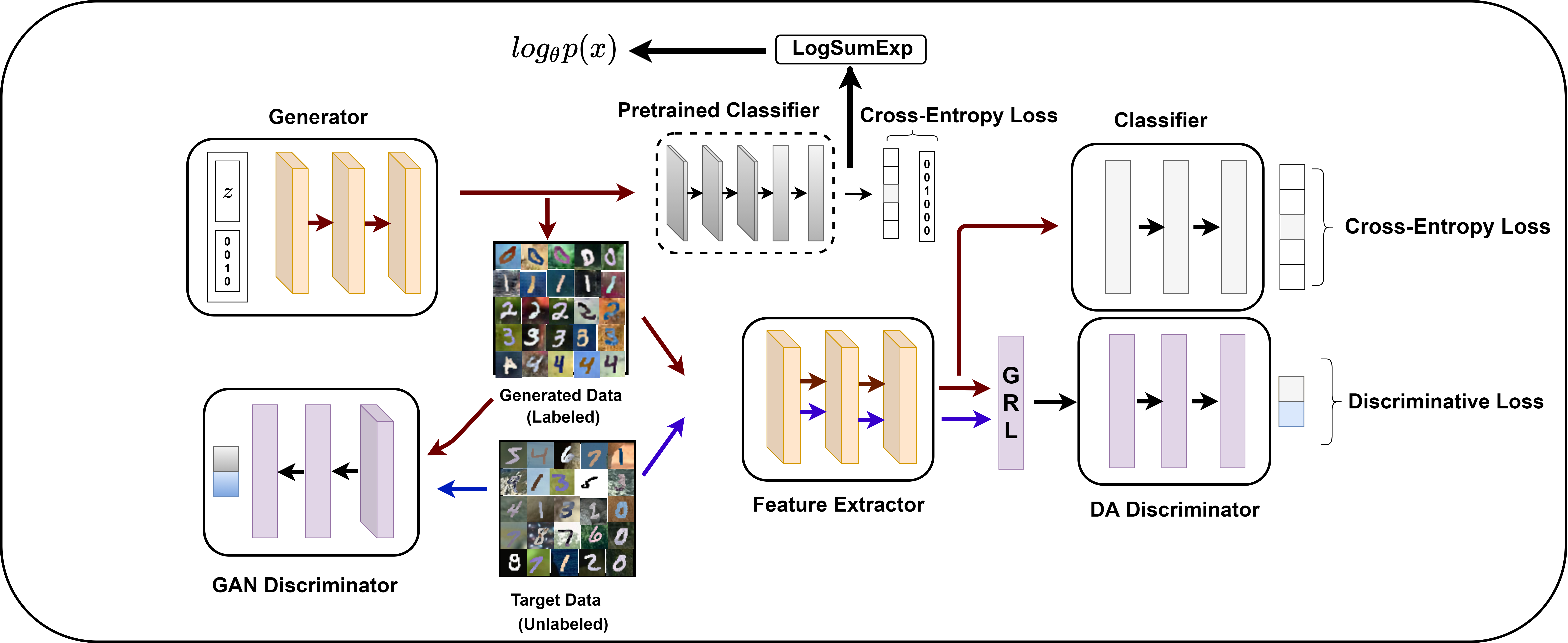}
    % \vspace{0.5em}
      \caption{Source Data free Domain Adaptation: Generator ($G$), GAN discriminator($D_{g}$), Feature extractor ($F$), Classifier ($C$) and Domain discriminator $(D_d)$ are trainable while the pre-trained Classifier ($P_c$) is set to frozen. $z$ is the latent noise vector. GRL is gradient reversal layer~\cite{ganin_ICML2015}.}
      \label{fig:main}
    %   \vspace{-1.5em}
 \end{figure*}

In this section, we discuss the source data free adaptation technique using a trained classifier. This problem is divided into two parts: the first part is to obtain the samples from the classifier, we call it the \textit{Generation module}. The second part is to adapt the classifier for the target domain, called \textit{Adaptation module}. These two modules are shown in Figure~\ref{fig:main}.

For the generation module, we work with the conditional GAN framework~\cite{mirza2014conditional} as a generative function to obtain the samples. The cross-entropy loss is used to obtain the domain impression and samples with class boundaries from the classifier. Note that by only the cross-entropy loss with GAN, we can enforce that generated samples follow only the conditional distribution $p(y|x)$. To learn the proxy samples of source data distribution, we model the joint distribution $p(x,y)$ defined in Eq.~\ref{eq:joint}.  For the adaptation module, we use the adversarial learning framework to make the feature invariant to the target domain with the generated data using a discriminator. 

\subsection{Problem Formulation}
The source data free domain adaptation problem can be formulated as follows. We consider a classifier $P_c$, which is trained on the source dataset $\mathcal{D}_s$ for the classification task. The assumption or constraint is that the source dataset is not available for adaptation. We are only provided the unlabeled target dataset $\mathcal{D}_t$ at training time. We further assume that the $\mathcal{D}_s$ comes from a source distribution $\mathcal{S}$ and $\mathcal{D}_t$ comes from a target distribution $\mathcal{T}$.  We assume that there are $N_t$ unlabeled target data points.

% In this work, we propose two variants of the source-free adaptation model. In the first variant, These variants of the proposed model have been discussed in detail in the subsequent sections.
\subsection{Proposed Model}
In the proposed method, we divide the model into two parts; one is a \textit{Generation} module, and the second one is an \textit{Adaptation} module.

\noindent
 \textbf{{Data Generation Module}}:
The proxy samples are obtained using a GAN framework with utilizing the source classifier.  The objective is to learn the joint distribution $p(x,y)$ of the source data. The basic idea behind this approach is to obtain the samples that can be perfectly classified by the classifier.  We use a parametric data generative neural network that is trained to maximize the log-likelihood defined in Eq.~\ref{eq:joint}. In this equation, the first term can be maximized using the derivative defined in Eq.~\ref{eq:derivate}.  The second term is optimized using the cross-entropy loss. A generative adversarial network, in conjunction with a trained classifier, is also applied to generate better samples. The vanilla GAN~\cite{radford2015unsupervised,goodfellow_NIPS2014} is an unconditional GAN and thus is not suitable here; because it is not guaranteed in the Vanilla GAN that only produce the specific desired class examples. So in the proposed generation framework, we use a conditional generative adversarial network~\cite{mirza2014conditional}, where the condition can be given as one-hot encoding and the latent noise vector to the generator to produce diverse samples. For obtaining the class-specific samples, we train this conditional generator with the cross-entropy loss of the classifier. In this case, we do not update the parameters of the pre-trained classifier; we only update the generator to produce the samples that can be classified as a given class vector. This formulation produces samples that may not be considered as natural samples, and it also produces adversarial noise examples. Thus these samples can not be used for further adaptation tasks. To obtain natural samples, we use an adversarial discriminator; it is trained with the help of target domain samples. The generator's parameters are updated with the adversarial loss from the discriminator and cross-entropy loss of the classifier.

% \begin{figure*}
%  \centering
%     \includegraphics[height=7.5cm,width=12.5cm]{fig/sf_gen.png}
%       \caption{ }
%       \label{fig:intro}
%       \vspace{-1.2em}
%  \end{figure*}
% \vspace{-0.30em}
\noindent\textbf{{Domain Adaptation Module:}}
The domain adaptation module consists of a shared feature extractor for source and target domain datasets, a classifier network, and a discriminator network similar to~\cite{ganin_ICML2015}. The discriminator's objective is to guide the feature extractor to produce domain invariant features using a gradient reversal layer. In the proposed framework, the domain discriminator is trained to discriminate between the generated labeled samples and the unlabeled target samples. Similarly, we fine-tune the trained classifier for the labeled generated samples. In this module, all networks, i.e., feature extractor, classifier, and discriminator, have learnable parameters. We also have experimented with the generation and adaptation processes separately. In this variant, we first train the generative model using the likelihood and GAN objective functions.  Then generative models parameters are set to be frozen and obtain samples. After that, these samples are used for adaptation. Here we have to fix the number of samples required for adaptation. The adaptation performance depends upon the number of the samples, as shown in the ablation study section in Table~\ref{tbl:results2}. 

\subsection{Loss functions}
The proposed Source Data free Domain Adaptation (SDDA) model is trained with these following losses.

\noindent\textbf{Likelihood based loss ($\mathcal{L}_{lik}$}): The objective is to learn a joint distribution of the source data from a discriminative model. This process required a maximize the log-likelihood of data obtained from the generative models as defined in Eq.~\ref{eq:joint}.
Thus loss function is written as 
% \vspace{-0.5em}
\begin{equation}
\mathcal{L}_{lik}=- \log p_{\theta}(\mathbf{x})
\end{equation}

\noindent The derivative of it is obtained from Eq.~\ref{eq:derivate}.

% \begin{equation}
% \frac{\partial\mathcal{L}_{lik}}{\partial \theta}=- \frac{\partial \log p_{\theta}(\mathbf{x})}{\partial \theta}
% \end{equation}

\noindent\textbf{Adversarial Loss ($\mathcal{L}_{adv}$}): This loss is used to train the GAN discriminator to discriminate between real data and data generated through the generator. The generator and GAN discriminator are adversaries. Here $a_i$ is a target data, sampled from $\mathcal{T}$, $y$ is the generated class label and $z$ is the latent noise vector, sampled from the normal distributions $P_z$.
% \begin{equation}
%      V(D_{g}, G)=\mathbb{E}_{\boldsymbol{a} \sim \mathcal{T}}
%      [\log D_{g}(\boldsymbol{a})]+
%      \mathbb{E}_{\boldsymbol{z} \sim P_{z}}[\log (1-D_{g}(G(\boldsymbol{z,y})))] 
%  \end{equation}
%  The objective function is $\min _{G} \max _{D_{g}} V(D_{g}, G)$.
 Loss for the  generator is defined as:
 \begin{equation}
      \mathcal{L}^g_{adv}= \sum_{i}\log (1-D_{g}(G(\boldsymbol{z_i,y_i})))
     \label{eq:adv}
 \end{equation}
% \vspace{-1em}
 Similarly loss for the GAN discriminator is defined as:
%  \vspace{-1em}
%  \begin{equation}
%       \mathcal{L}^d_{adv}= - (\sum_{\boldsymbol{a} \sim \mathcal{T}} \log D_{g}(\boldsymbol{a_i})+  \sum_{i}\log (1-D_{g}(G(\boldsymbol{z_i,y_i}))))
%      \label{eq:advd}
%  \end{equation}

 \begin{equation}
      \mathcal{L}^d_{adv}= (\sum_{i} \log D_{g}(G(\boldsymbol{z_i,y_i}))+  \sum_{\boldsymbol{a}_i \sim \mathcal{T}}\log (1-D_{g}(\boldsymbol{a_i})))
     \label{eq:ad_dis}
 \end{equation}

%  \vspace{-3em}
%  \begin{equation}
%      \mathcal{L}_{adv}=\min _{G} \max _{D_{g}} V(D_{g}, G)
%      \label{eq:adv}
%  \end{equation}

\noindent\textbf{Cross-Entropy Loss ($\mathcal{L}_{crs}$)}: This loss is obtained by passing the generated images to the pre-trained classifier. The predicted output of the pre-trained classifier is compared with the class vector that is input to the generator. This loss does not update the parameters of  pre-trained classifier. It only updates the parameters of the generator to produce class consistent images.
% \vspace{-0.5em}
\begin{equation}
     \mathcal{C}_{crs}=\frac{1}{N_g}\sum_{g_i \in \mathcal{D}_g} \mathcal{L}_c(P_c(g_i)),y_i) 
     \label{eq:crs}
\end{equation}

% \begin{equation}
%      \mathcal{L}_{crs}=\min _{G} \mathcal{C}_{crs}(G)
%      \label{eq:crs}
% \end{equation}
% \vspace{-1em}
Where  $g_i=G({z_i,y_i})$ is a generated image sample. $\mathcal{L}_c$ is the tradition cross entropy loss. $N_g$ are the generated samples. $P_c$ is the pre-trained classifier.

\begin{table*}[!]
\centering
\renewcommand{\arraystretch}{1.1} % Default value: 1
 \scalebox{0.9}{
\begin{tabular}{|c|c|c|c|c|c|c|}
 \hline
 \textit{{Source Data Required}} &\textit{Method }& \textit{MNIST$\rightarrow$MNIST-M} & \textit{SVHN$\rightarrow$MNIST }&  \textit{MNIST$\rightarrow$SVHN}  &   \textit{MNIST$\rightarrow$USPS} \\ 
  \hline
  \hline
 & DANN~\cite{ganin_ICML2015}  &  81.5 &71.1 & 35.7 &89.1 \\
   & CMD~\cite{zellinger2016central}  & 85.5 &86.5 & - &   \\
    &  kNN-Ad~\cite{sener2016learning}  & 86.7 &78.8 & 40.3 &  \\ 
      & DRCN~\cite{ghifary_ECCV2016}  & - & 82.0 & 40.1 & 91.8  \\
 \textbf{Yes}     & PixelDA\cite{bousmalis_2017CVPR} & 98.2 &- & -  &  \\
       & ADDA~\cite{tzeng_CVPR2017}  & - &76.0 & -  &  \\
       &  ATN~\cite{saito2017asymmetric}  & 94.2 &86.2& 52.8 & -  \\
 & MCD\cite{saito2018maximum} & - & 96.2 & -  &  \\

 & JDDA\cite{chen2019joint}& 88.4 &94.2 & - &   \\
 & UDA~\cite{cicek2019unsupervised} & 99.5 &99.3 &89.2  &  \\
 & 3CATN~\cite{li2019cycle} &  - & 98.3 & - & 96.1 \\
\hline
 & Baseline & 59.4 & 67.2 &  37.7  & 82.5  \\ \cline{2-6}
%  \multirow{2}{*}{}  & On Source Data & 87.5 & - & -& - \\ 
  \multirow{2}{*}{} \textbf{No} & SDDA(ours) & \textbf{85.5} & 75.5 & 42.2 &  \textbf{89.9} \\ 
 & SDDA-P(ours) & 84.1 & \textbf{76.3} & \textbf{43.6}  & 88.5 \\  \hline
%  \hline
 
% Tested on Source Data  & SDDA-P(ours) & 87.5 & 97.8 & 84.6& 95.3 \\  \hline
%
% \textbf{Yes} & Oracle & 82.5 & 75.3 & 39.8 \\
%\hline
\end{tabular}
}

\caption {Classification accuracy (\%) comparisons with baseline and other state-of-the-art methods on standard digit dataset using the proposed method. Note that the proposed models do not use the source samples for adaptation, while all state-of-the-art methods access the source data. The baseline is without the adaptation method.  SDDA-P  is referred to when we initialized the classifier with the weight of a pre-trained classifier.\label{tbl:results}} 

% \label{tbl:imageclef_res}
% \end{center}
\end{table*}

\begin{table*}[!]
\centering
\renewcommand{\arraystretch}{1.1} % Default value: 1
 \scalebox{0.9}{
\begin{tabular}{|c|c|c|c|c|c|c|c|c|c|}
 \hline
 \textit{{Source Data Required}} &\textit{Method }& \textit{A$\rightarrow$W} & \textit{D$\rightarrow$W}&  \textit{W$\rightarrow$D}  &   \textit{A$\rightarrow$D} & \textit{D$\rightarrow$A}& \textit{W$\rightarrow$A}&  \textit{Avg} \\ 
  \hline
 & DANN~\cite{ganin_ICML2015} &  81.2 & 98.0 & 99.8 & 83.3 &  66.8 & 66.1 & 82.5 \\
 
  \textbf{Yes} & GTA~\cite{sankaranarayanan2018generate}  &  89.5 & 97.9 & 97.9 & 87.7 &72.8 &  71.4 & 86.5 \\
  
 & DADA~\cite{tang2020discriminative} &  92.3& 99.2& 100.& 93.9& 74.4& 74.2& 89.0 \\
\hline
\hline
 & Baseline & 79.9 &  96.8  & 99.5 &  84.1& 64.5 &  66.4&  81.9  \\ \cline{2-9}
  \multirow{2}{*}{} \textbf{No} & SDDA(ours) & \textbf{82.5} & \textbf{99.0}&  \textbf{99.8}& \textbf{85.3}&\textbf{66.4} &  \textbf{67.7} & \textbf{83.5} \\ 
 \hline
%
% \textbf{Yes} & Oracle & 82.5 & 75.3 & 39.8 \\
%\hline
\end{tabular}
}

\caption {Classification accuracy (\%) comparisons with baseline and other state-of-the-art methods on Office-31~\cite{saenko_ECCV2010} dataset using proposed method. Note that, the proposed models do not use source samples for adaptation, while all other methods utilize the source data for adaptation. \label{tbl:results_office31}} 
%\vspace{-2em}

% \label{tbl:imageclef_res}
% \end{center}
\end{table*}

\begin{table}[h]
\centering
\renewcommand{\arraystretch}{1.1} % Default value: 1
%\vspace{-2em}
 \scalebox{0.8}{
\begin{tabular}{|c|c|}
 \hline
 \textit{{\textbf{Dataset}}} &\textit{\textbf{Performance}}
  \\ \hline
  \textit{MNIST$\rightarrow$MNIST-M } & 87.5
  \\  \hline
\textit{SVHN$\rightarrow$MNIST } & 97.8
  \\  \hline
 \textit{MNIST$\rightarrow$SVHN}  & 84.6  \\  \hline   \textit{MNIST$\rightarrow$USPS}  &  95.3
 \\  \hline

\end{tabular}
}

\caption {Classification performance on source data after the adaptation.\label{tbl:cls_results}} 
%\vspace{-2em}
% \label{tbl:imageclef_res}
% \end{center}
\end{table}

\noindent\textbf{Domain Discriminative Loss ($\mathcal{L}_{dis}$)}: This loss is used to obtain domain invariant features from the feature extractor. It is a binary classification loss between the source and target samples. The discriminator is trained with the gradient of loss. In contrast, the feature extractor is trained by the negative gradient of this loss (using gradient reversal layer~\cite{ganin_ICML2015}) to obtain domain invariant.
\begin{equation}
   \mathcal{L}_{dis}= \frac{1}{N} \sum_{x_i \in \mathcal{D}_g \cup \mathcal{D}_t} \mathcal{L}_c(D_{d}(F(x_i)),d_i)
   \label{eq:dis}
\end{equation}
% \begin{equation}
%     \mathcal{L}_{dis}=\min _{D_d} \mathcal{C}_{dis} +\max _{F} \mathcal{C}_{dis}
%     \label{eq:dis}
% \end{equation}
$N$ is the total number of generated and target samples. $d_i$ is the domain label, where $ d_i=0$ $\text{if $x_i \in \mathcal{D}_g $}$ and $ d_i=1$ $\text{if $x_i \in \mathcal{D}_t $}$. $\mathcal{L}_c$ is the normal cross-entropy loss.

\noindent\textbf{Classification Loss ($\mathcal{L}_{cls}$)}: The adaptive classifier is trained using the classification loss of generated samples. We update this classifier's parameters based on the loss gradient. The gradient of this loss is also used to train feature extractor to generate class discriminative features.
% \vspace{-0.5em}

\begin{equation}
    \mathcal{L}_{cls}= \frac{1}{N_g}\sum_{g_i \in \mathcal{D}_g} \mathcal{L}_c(C(F(g_i)),y_i) 
    \label{eq:cls}
\end{equation}
% \begin{equation}
%      \mathcal{L}_{cls}= \min _{C,F}  \mathcal{C}_{cls}
%      \label{eq:cls}
% \end{equation}
Here $C$ is the classifier network. $N_g$ are the total number of generated samples.

\noindent\textbf{Total Loss:}
The total loss is given as below
\begin{equation}
    \mathcal{L}(G,F,D_d)= \delta\mathcal{L}_{lik} +\alpha*\mathcal{L}^g_{adv} + \beta*\mathcal{L}_{crs}+ \lambda*\mathcal{L}_{dis}+ \mu*\mathcal{L}_{cls}
\end{equation}

% \begin{equation}
%     \mathcal{L}^d(D_g)=\mathcal{L}^d_{adv}
% \end{equation}
where $ \delta, \alpha$, $\beta$, $\lambda$ and $\mu$ are the tuning parameters. In our experiments, $\alpha$ and $\beta$ are set to 1 and exponentially decreased to 0 while $\mu$ is kept 0 until 25 epochs, and later it is set to 1.  $\lambda$ is the adaptation parameter. It is set to 1 throughout the experiments. we set $\delta=0.1$ in all the experiments.
We also optimize the parameters of the adversarial discriminator by minimizing the loss defined in Eq.~\ref{eq:ad_dis} for a given generator's parameters.

\section{Results and Discussion}

% \begin{figure}[h]
% \begin{subfigure}{0.3\textwidth}
%   \centering
%     \includegraphics[scale=0.4]{proxy_new.png}
%   \caption{}
%   \label{fig:Proxy}
% \end{subfigure}
% \begin{subfigure}{0.3\textwidth}
%   \centering
%     \includegraphics[scale=0.17]{ssa.png}
 
%   \caption{}
%   \label{fig:ssa}
% \end{subfigure}
%     %   \vspace{-0.7em}   
% \caption{ (a) Proxy distance between source-target, source-generated and target-generated domains before and after the adaptation for MNIST$\rightarrow$ MNIST-M adaptation task. In this figure the S, T and G stands for source data, target data and generated data respectively. (b) SSA plots for SDDA, source only and oracle model with a significance level of 0.05. The mean rank is plotted on x-axis. The CD calculated as 0.6051 and all the methods are way outside the CD, are statistically different. Here SDDA is proposed source data free domain adaptation model, Oracle is when source data is used and Source only is without adaptation model for MNIST$\rightarrow$ MNIST-M adaptation task}  
% \label{fig:intro_fig}
% \end{figure}

\subsection{Datasets}
\textbf{MNIST$\rightarrow$ MNIST-M:}
We experiment with the MNIST dataset~\cite{lecun1998gradient} as source
data. In order to obtain the target domain (MNIST-M)
we blend digits from the original set over patches randomly extracted from color photos from BSDS500~\cite{arbelaez2010contour}. Due to this, a domain gap is observed, and performance is poor on the MNIST-M classifier. There are 60k samples used to train the MNIST classifier, and 59k samples of MNIST-M are used for adaptation. For adaptation results are shown in Table~\ref{tbl:results}.
% \vspace{-1em}

\textbf{SVHN$\rightarrow$ MNIST:}
In this adaptation task, source data (SVHN~\cite{netzerreading}) and target data (MNIST) both have ten-classes. In this setting, we are provided the classifier trained on the SVHN and unlabeled MNIST dataset. The provided classifier is trained on the full SVHN dataset, and we adapt the full MNIST dataset. There are 60k samples present in the MNIST dataset, while SVHN has ~73K samples. The results are reported in Table~\ref{tbl:results}.

\textbf{MNIST$\rightarrow$ SVHN:}
For the MNIST-SVHN transfer task, the provided pre-trained classifier trained on the MNIST dataset. We use the full SVHN dataset. The classifier is also trained on the full MNIST dataset. The results are reported in Table~\ref{tbl:results}.
% \vspace{-1em}

\textbf{MNIST$\rightarrow$ USPS:}
The USPS contains 16x16 grey images. We resized them to 32x32. In this experiment, we use full MNIST and USPS images as the target set.  The results are reported in Table~\ref{tbl:results}.

\textbf{Office-31~\cite{saenko_ECCV2010}:}
It contains three domains Amazon (A), Webcam (W), and DSLR (D). Each domain has 31 object classes, and we evaluate all the six adaptation task. We obtain the features from ResNet-50~\cite{he2016deep}, pre-trained on Imagenet.

\subsection{Performance Evaluation}
Table~\ref{tbl:results} shows the results for different adaptation tasks for the proposed method. In the table, baseline refers to the case when there is no adaptation performed. This is one of the pioneering efforts to solve domain adaptation without accessing source data to the best of our knowledge. The SDDA-P is referred to when the classifier is initialized with the pre-trained classifier weight, while SDDA is when it is initialized randomly. Note that all the previous state-of-the-art methods work when source data is accessible. Table~\ref{tbl:results} shows that the proposed model performs comparably to the baselines that make use of full source information.
Table~\ref{tbl:cls_results} shows the classifier's performance on  source dataset after the adaptation. In target data, we achieve a boost in the performance from the baselines; for example, from MNIST $\rightarrow$MNIST-M adaptation task, we obtained $\sim 25$\% improvement. For the other adaptation task, we also obtained improvement with a large margin. In Table~\ref{tbl:results2}, the results for the other variant in the training method, we call it adaptation after the generation (SDDA-G), are presented for the MNIST$\rightarrow$MNIST-M and MNIST$\rightarrow$SVHN adaptation tasks. In this method, we first learn the generative model, and after that, samples are generated to train the adaptation module. The number in the bracket indicates the number of generated samples used for the adaptation. We can observe that initially, the performance improves when we increase the number for generated samples, but later it slightly deteriorates.

For Office-31 dataset, the adaptation results on all the six tasks are reported in Table~\ref{tbl:results_office31}.  In the dataset adaptation, we generate the features of corresponding images from the generator. We can observe that we can achieve the  $\sim$ 3\%  and  $\sim$ 1.5 \% improvement over the baseline without accessing the source dataset on hard adaptation task A$\rightarrow$W and  A$\rightarrow$D. We implement on Torch-Lua framework. 
% The  other training details are provided in the supplementary materials.

\begin{table}
% \vspace{-1em}
\centering
\renewcommand{\arraystretch}{1.05} % Default value: 1

%  \begin{center}[ht]
 \scalebox{0.8}{
\begin{tabular}{|c|c|c|}
 \hline
  {\textit{Method} }&\textit{ MNIST$\rightarrow$MNIST-M}  & \textit{MNIST$\rightarrow$SVHN } \\ 
  \hline
Source Only(0 samples) & 59.4 & 37.7  \\
\hline
SDDA-G(300 samples)) & 64.3 & 38.5 \\
\hline
SDDA-G(2000 samples)) & 61.8 & \textbf{39.6} \\
\hline
SDDA-G(6000 samples)) & \textbf{70.5} & 38.8 \\
\hline
SDDA-G(40000 samples)) & 68.7 & 39.2\\
\hline
\hline
Oracle & 82.5 & 39.8 \\
\hline

\end{tabular}
}
\caption {Classification accuracies for MNIST$\rightarrow$MNIST-M and  MNIST$\rightarrow$SVHN  transfer task for different generated samples for \textit{Adaptation after Generation} variant.
Our model is SDDA-G with the number in bracket indicating the number of generated  samples used for adaptation. Oracle refer, when actual source data is used for adaptation.} 
\label{tbl:results2}
% \end{center}
\end{table}

\section{Analysis}

\subsection{Ablation study on Loss Functions}
In Table~\ref{tbl:loss}, we show the ablation study of different loss functions used by the proposed model for the MNIST $\rightarrow$MNIST-M adaptation task. We can observe that by introducing the likelihood-based loss, we get better improvement. The generative adversarial loss is very crucial to incorporate; the model does not converge without it. The reason is that the generator can not be trained without any adversarial discriminator.
\begin{table}
\begin{tabular}{|c|c|c|c|c|c|}
\hline
$\mathcal{L}_{lik}$ & $\mathcal{L}_{dis}$  & $\mathcal{L}^g_{adv} $ & $\mathcal{L}_{crs}$  & $\mathcal{L}_{cls}$  & \textbf{Accuracy} \\
\hline
- & - & $ \surd$ & $\surd$ & $\surd$ & 80.6 \\
\hline
-  & $ \surd$ & $ \surd$ & $\surd$ & $\surd$ & 83.1 \\
\hline
$\surd $ & $ \surd$ & - & $\surd$ & $\surd$ & not converged \\
\hline
$\surd $ & $ \surd$ & $ \surd$ & $\surd$ & $\surd$ & \textbf{85.5} \\
\hline

\end{tabular}
\caption{Ablation study of different loss functions for the MNIST$\rightarrow$MNIST-M adaptation task.}
\label{tbl:loss}
%\vspace{-2em}
\end{table}

\begin{figure*}[!]
 \centering
    \includegraphics[height=8.0cm,width=13.5cm]{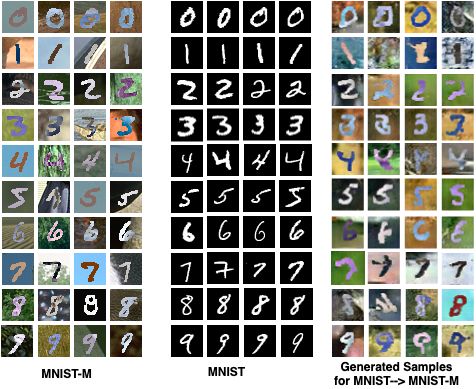}
       \caption{Visualization of source data (MNIST), Target data (MNIST-M)and Generated data for class digit 0-9. We can observe that generated images have proper class discrimination. }
      \label{fig:vis}
      %\vspace{-1em}
 \end{figure*}

\subsection{Ablation on other Domain Adaptation models}
This section provides the results for different domain adaptation methods such as MMD~\cite{tzeng_arxiv2014}, IDDA~\cite{kurmi2019looking}, Wasserstein DA~\cite{shen_AAAI2018wasserstein} and GRL~\cite{ganin_ICML2015}. In these experiments, we use DCGAN architecture for both generator and classifier. This analysis reveals that the proposed method can be plugged into any domain adaptation framework. Results are shown in Table~\ref{tbl:abl} for the MNIST$\rightarrow$MNIST-M adaptation task
% &  &  &  &   & -

\subsection{Distribution Discrepancy}
\begin{figure}[!h]
  \centering
    \includegraphics[scale=0.35]{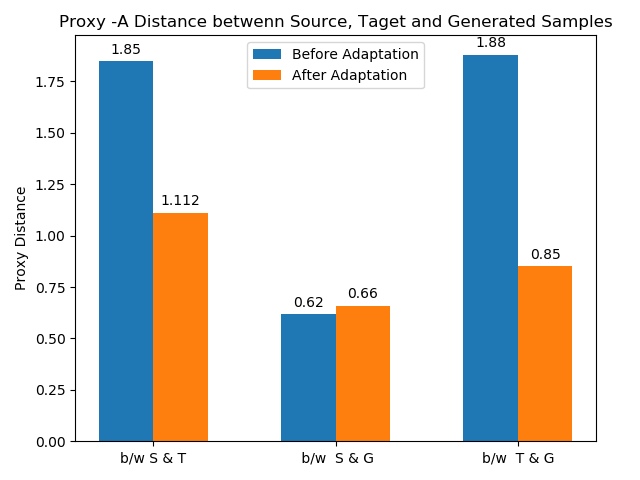}
 
\caption{ Proxy distance between source-target (S\&T), source-generated (S\&G) and target-generated (T\&G) domains before and after the adaptation for MNIST$\rightarrow$ MNIST-M adaptation task. In figure, S, T, and G stands for source data, target data, and generated data respectively.}
 \label{fig:Proxy}
\end{figure}

 The domain adaptation theory~\cite{ben_ML2010} suggests $\mathcal{A}$-distance as a measure of a cross-domain discrepancy, which, together with the source risk,  bounds the target risk. The proxy $\mathcal{A}$-distance is defined similar to~\cite{pei_arxiv2018} as  $d_{\mathcal{A}} = 2(1-2\epsilon)$,
where $\epsilon$ is the generalization error of a classifier (e.g. kernel
SVM) trained on the binary task of discriminating source and
target. Figure~\ref{fig:Proxy} shows $d_{\mathcal{A}}$ for MNIST$\rightarrow$ MNIST-M adaptation task between source-target, source-generated,and target-generated domains in before adaption and after the adaptation. We can infer from the figure that source and generated domains are always closer. The target domain is closer after the adaptation as compare to before adaptation model.

\begin{table}[t]
\scalebox{0.8}{
\begin{tabular}{|c|c|c|c|}
 \hline
 \textbf{  Method} &\textbf{ Source Only} & \textbf{Full Source data} & \textbf{ Source free(Proposed)} \\
 \hline
 MMD~\cite{tzeng_arxiv2014} & 59.1 & 64.3 &  62.5  \\
 \hline
 IDDA~\cite{kurmi2019looking}   & 59.1  &  82.3 & 83.0   \\
  \hline
 WDA~\cite{shen_AAAI2018wasserstein}& 59.1  & 82.8  &  79.5  \\
  \hline
 GRL~\cite{ganin_ICML2015}& 59.1  & 82.5 & \textbf{85.5}   \\
  \hline
\end{tabular}
}
\caption {Performance of different domain adaptation model on MNIST$\rightarrow$MNIST-M adaptation task. Source only: when there is no-adaptation, Full source data: full source data is used for adaptation, the Proposed method: samples are generated from the trained classifier, and for adaptation, these dummy samples are used. } 
\label{tbl:abl}
% \vspace{-1em}
\end{table}

\begin{table}[!h]
\renewcommand{\arraystretch}{1.05} % Default value: 1
\centering
\scalebox{0.8}{
%  \begin{center}[ht]
\begin{tabular}{c|c|c|c|c|c|c|c}
 \hline
  \textbf{$\lambda$}  &$0.1$ &$0.3$&$0.5$&$0.8$ &$1$ &$1.5$ &$2$\\ 
  \hline
    SDDA-G(6k) &  66.3 & 66.5 & 68.7  & \textbf{70.8} & 70.5 &66.9  &- \\
  \hline
     SDDA-P& 78.9 & 80.2 & {84.0}  & 84.0 & \textbf{84.1} &79.3  &77.8 \\
  \hline
    SDDA &  82.8 & 83.4 & 83.5  & 83.2 & \textbf{85.5} & 82.6  &82.3 \\
  \hline
\end{tabular}
}
\caption {Ablation study of adaptation parameter $\lambda$ for MNIST$\rightarrow$MNIST-M adaptation task.} 

\label{tbl:lamda}
% \end{center}
% \vspace{-1em}
\end{table}

\begin{table}[]
\centering
\scalebox{0.9}{
\begin{tabular}{c|c|c|c|c}
\hline
 & \multicolumn{2}{c|}{No Adapt} & \multicolumn{2}{c}{Adapt} \\ \hline
 &   around Src &  around Tgt      &   around Src &  around Tgt    \\ \hline
  Density &     0.771     &   0.769      &    0.734      &  0.776       \\ \hline
\end{tabular}
}
\caption {Density estimation  generated samples around the source (\textbf{Src}) and target (\textbf{Tgt}) domains for MNIST$\rightarrow$MNIST-M adaptation task on adapted and non-adapted features.} 
\label{tbl:density}
%\vspace{-2em}
\end{table}

\subsection{Performance on number of Generated Samples}
We experiment on the number of generated samples required for the adaptation. In this setting, we first generate the samples without the adaptation module. Both generation and adaptation modules are trained separately. We have experimented with this variant in the MNIST$\rightarrow$ MNIST-M  and MNIST $\rightarrow$ SVHN adaptation tasks with different numbers of generated samples. The performance is reported in Table~\ref{tbl:results2}. From the table, we can observe that we obtain the best performance MNIST$\rightarrow$ MNIST-M when the number of generated samples is selected 6000.

\subsection{Image Generation Visualization }
In Figure~\ref{fig:vis}, we provide visualization of  generated imaged during the adaptation process for MNINT$\rightarrow$MNIST-M adaptation task. Here MNIST data is source data, and MNIST-M data is target data. We can observe that the generated images look like the target samples. The reason is that we use the target samples as real data for training the GAN. We can also observe that the generated images are class discriminative, i.e., each sample has one class. This implies that the cross-entropy loss from the pre-trained classifier helps  generator to provide the class structure so it can avoid the mode collapsed problem. The third observation is that all the examples are diverse so that we can generate sufficient distinct examples to train the classifier.

% \begin{table}[!h]
% \renewcommand{\arraystretch}{1.05} % Default value: 1
% \centering
% %  \begin{center}[ht]
% \begin{tabular}{c|c|c}
%  \hline
%   & Around Source & Around Target \\
%   \hline
  
%  Density & 0.73 & 0.78 \\
%  Average Distance & 182.73 & 45.32 \\
 
%   \hline
% \end{tabular}
% \caption {Density estimation and average distance of generated samples arround the source and target domains for MNIST$\rightarrow$MNIST-M adaptation task.} 
% \label{tbl:density}
% \end{table}

\subsection{Ablation Study with Adaptation parameter $\lambda$}
We provide ablation of proposed method for value of $\lambda$ in Table~\ref{tbl:lamda} for MNIST$\rightarrow$MNIST-M adaptation. It can be observed in the adaption; the proposed model is not very sensitive to the adaptation value. Performance is better when we choose $\lambda=1$ for the SDDA model.

. 
\subsection{Density Estimation of Generated samples}
 The objective of density estimation is to estimate the closeness of generated samples with source and target domains. We estimate density in both cases i.e around the source data and the target data~\cite{jamal2018u}. For obtaining it, the features are obtained by forward images till convolution layers.  We analyze the density estimation using both adapted and non-adapted features. The generated samples density around the source domain is the average number of samples, which can be found within a $\epsilon$ neighborhood of source samples.  These results are reported in Table~\ref{tbl:density}. This density estimation shows that the generated samples have a similar density with source and target data using the non-adapted model for features. It shows that the distribution of generated samples is equally close to both the source and target dataset.  However, in adapted features, the generated samples' density is slightly higher around the target domain. 

 \section{Conclusion}
%  \vspace{-1em}
We propose a source data-free adaptation method that solves one of the critical challenges that existing domain adaptation techniques face, i.e., the availability of source data. The proposed approach is generic, i.e., it can be applied with any existing domain adaptation models. The proposed work is one of the novel attempt that tackles the domain adaptation problems without the source data's availability. From the results obtained, we believe that the proposed model provides an exciting avenue for further research on this problem.

{\small
\bibliographystyle{ieee_fullname}
\bibliography{egbib}
}

\end{document}